\documentclass[runningheads]{llncs}

\usepackage[utf8]{inputenc}
\usepackage{graphicx}
\usepackage{amsmath}

\usepackage[english]{babel}

\usepackage[utf8]{inputenx}
\usepackage[T1]{fontenc}
\usepackage{lmodern}
	
\usepackage{color, colortbl}
\usepackage{tabularx}
\usepackage{booktabs}

\usepackage{amssymb}
\usepackage{amsmath}  % https://tex.stackexchange.com/a/715003
\usepackage{amsfonts}
\usepackage{amscd}

\usepackage{mfirstuc}
\usepackage{dirtytalk}

\usepackage{xcolor}

\usepackage{makecell, multirow}

\usepackage{float}  % https://tex.stackexchange.com/a/8633/226150

\usepackage{subcaption}
\usepackage{cite}

\usepackage[title]{appendix}
\usepackage{url}
\usepackage{hyperref}

\usepackage{epigraph}
\usepackage{prettyref}
\newrefformat{ch}{chapter~\ref{#1}}  % on page \pageref{#1}}
\newrefformat{sec}{section~\ref{#1}}  % on page \pageref{#1}}
\newrefformat{subsec}{Section~\ref{#1}}  % on page \pageref{#1}}
\newrefformat{eq}{Equation~\ref{#1}}  % on page \pageref{#1}}
\newrefformat{thm}{Theorem~\ref{#1}}  % on page \pageref{#1}}
\newrefformat{fig}{Figure~\ref{#1}}  % on page \pageref{#1}}
\newrefformat{tab}{Table~\ref{#1}}  % on page \pageref{#1}}
\newrefformat{alg}{Algorithm~\ref{#1}}  % on page \pageref{#1}}

\newcommand{\klinik}[1]{local University Hospital}  % anonym due to double-blind, later: Innsbruck University Hospital
\newcommand{\si}[1]{the Supplementary Information}
\newcommand{\minisection}[1]{\noindent{}\textbf{#1.}}

\newcommand{\pvalSign}[1]{$p$-value significance is calculated with respect to the \textit{baseline}}
\definecolor{gray}{rgb}{0.2, 0.2, 0.2}

\newcommand{\makespace}{\quad~\quad}
\title{Implicit Deformable Medical Image Registration with Learnable Kernels}
\author{Stefano Fogarollo\inst{1}, Gregor Laimer\inst{2}, Reto Bale\inst{2}, Matthias Harders\inst{1}}  % Anonymous

\institute{Department of Computer Science Interactive Graphics and Simulation Group (IGS), University of Innsbruck, Technikerstraße 21 Innsbruck, Austria \and Interventional Oncology-Microinvasive Therapy (SIP), Department of Radiology, Medical University Innsbruck, Innsbruck, Austria}
\authorrunning{Stefano Fogarollo, Gregor Laimer, Reto Bale, Matthias Harders}

\begin{document}

\maketitle

\begin{abstract}

Deformable medical image registration is an essential task in computer-assisted interventions. This problem is particularly relevant to oncological treatments, where precise image alignment is necessary for tracking tumor growth, assessing treatment response, and ensuring accurate delivery of therapies. Recent AI methods can outperform traditional techniques in accuracy and speed, yet they often produce unreliable deformations that limit their clinical adoption. In this work, we address this challenge and introduce a novel implicit registration framework that can predict accurate and reliable deformations. Our insight is to reformulate image registration as a signal reconstruction problem: we learn a kernel function that can recover the dense displacement field from sparse keypoint correspondences. We integrate our method in a novel hierarchical architecture, and estimate the displacement field in a coarse-to-fine manner. Our formulation also allows for efficient refinement at test time, permitting clinicians to easily adjust registrations when needed. We validate our method on challenging intra-patient thoracic and abdominal zero-shot registration tasks, using public and internal datasets from the \klinik{}. Our method not only shows competitive accuracy to state-of-the-art approaches, but also bridges the generalization gap between implicit and explicit registration techniques. In particular, our method generates deformations that better preserve anatomical relationships and matches the performance of specialized commercial systems, underscoring its potential for clinical adoption.

\end{abstract}

\section{Introduction}

\begin{figure}[!h]
\centering
\includegraphics[width=\textwidth]{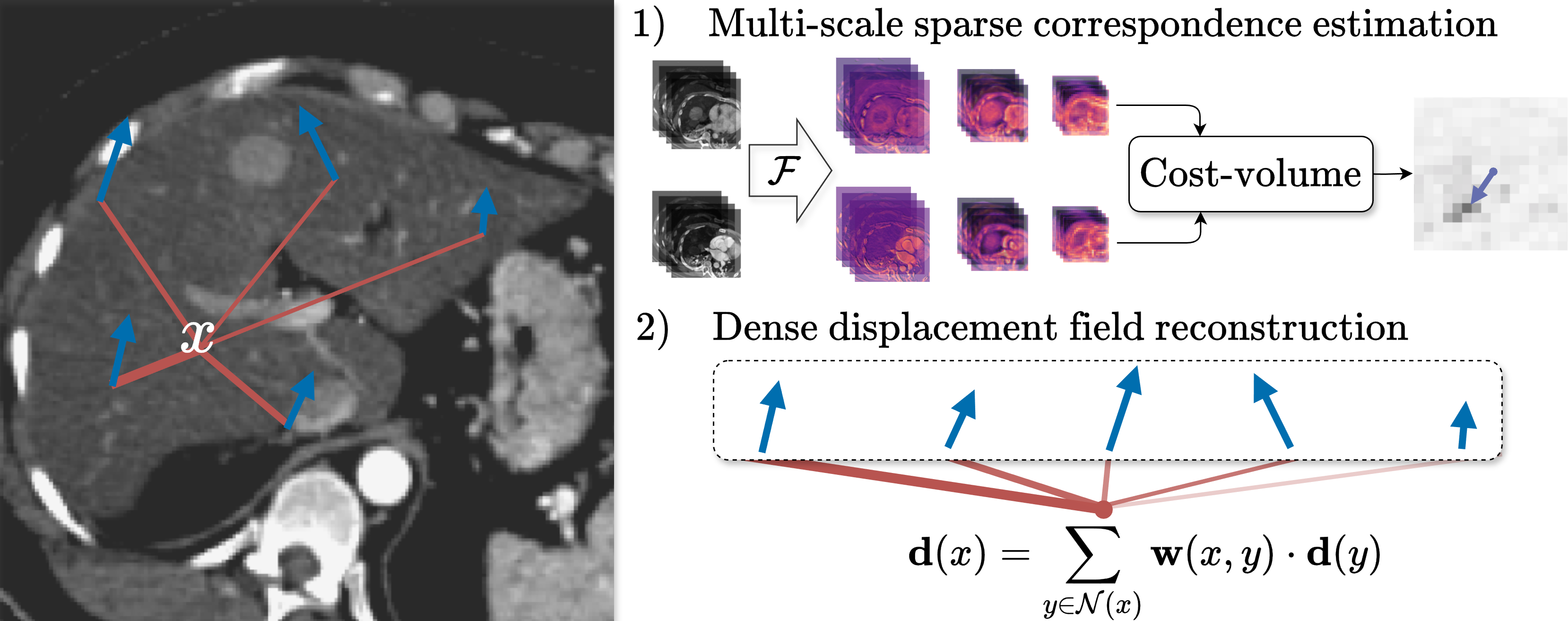}
\caption{In this work, we rephrase deformable image registration as signal reconstruction. First, we obtain sparse correspondences between the input images using cost-volume optimization on multi-scale features learned with $\mathcal{F}$. Then, we reconstruct the displacement field with a learnable kernel function $\mathbf{w}$, conditioned on the local displacements $\mathbf{d}(y), y \in \mathcal{N}(x)$, depicted in blue.}
\label{fig:method}
\end{figure}

Accurate and reliable registration is an essential step in computer-assisted interventions, with direct applications on intra-procedural navigation, treatment monitoring and evaluation. The task consists of finding the optimal transformation that aligns the two input images. In particular in the abdominal and thoracic regions, deformable registration is needed to correctly model large nonlinear deformations resulting from the complex behavior and interaction of soft tissues \cite{liver_motion_caused_by_ventilation}.
Traditional image registration methods \cite{Syn, Elastix, modat_registration} can estimate accurate deformations for each image pair, but often struggle to balance accuracy, computational efficiency, and anatomical plausibility; modern techniques can leverage hardware acceleration (GPU) for faster inference \cite{convexadam, iterlbp}.
Recent advances in artificial intelligence (AI) have shown promising results in addressing these challenges, with learning-based methods achieving state-of-the-art performance in terms of both speed and accuracy. Building on the frameworks described in VoxelMorph \cite{VoxelMorph} and SynthMorph \cite{synthmorph}, AI methods are trained to predict a tensor $\phi \in \mathbb{R}^{D \times H \times W \times 3}$ representing the displacement field that spatially aligns the two $3$D inputs, namely the moving and fixed images, with $D, H, W$ denoting the spatial dimensions. Backpropagation of the gradients is achieved by warping the image with $\phi$ using a Spatial Transformer Network layer \cite{stn}. Multi-stage incremental prediction has been shown to increase the registration accuracy with minimal computational overhead \cite{nicetrans, modet, winet, rdp}, achieving comparable performance to pyramidal \cite{lapirn} and cascaded \cite{zhao_recursive_2019} architectures. % Mention nerf and diffusion? But these are not architectures ..
Despite these improvements, current AI registration methods struggle with unseen anatomical variations or clinical scenarios not encountered during training \cite{learn2reg}. Recent foundation models try to bridge this gap by pre-training on large-scale diverse datasets, but they still need test-time refinement in difficult zero-shot cases (``Type 2'' out-of-distribution) \cite{unigradicon, MultiGradICON}. Related to our approach are implicit neural representations (INRs), which employ multi-layer perceptrons (MLP) to continuously represent a signal \cite{siren_inr, modulated_periodic}. While INRs offer significant potential in medical imaging due to their inherent efficiency \cite{nerf}, they are designed to fit individual input signals, rather than to generalize to unseen data. The implicit method introduced in \cite{implicit_reg} can model the spatial transformation between a pair of medical images, but requires retraining for each new pair. Several strategies have been explored to address this limitation, including hyper-networks \cite{siren_inr, trans_meta_learners, Feh_Intraoperative_MICCAI2024}, modulation of periodic activations \cite{modulated_periodic}, and conditioned MLPs \cite{pmlr-v172-amiranashvili22a}. Recently, the authors of \cite{towards_gener_implicit_reg} propose to generalize the pairwise implicit registration technique of \cite{implicit_reg} by conditioning on learned image features. To the best of our knowledge, this is the only prior work addressing generalized INRs for medical image registration, highlighting a critical research gap. Therefore, a key question remains \cite{survey_registration_mia25}: \textbf{how can we mitigate AI generalization issues to achieve reliable zero-shot registration?} This is the focus of our work, and to answer the question, we cast deformable image registration as a signal reconstruction problem. We propose to reconstruct the dense displacement field with a learnable kernel conditioned on sparse keypoint correspondences. An implicit dual-stream attention mechanism is used to model the spatial and semantic dependencies of the neighboring displacements, allowing to learn complex deformations in a data-driven manner. Distinct from sparse correspondence extrapolation methods \cite{GraphRegNet}, our approach learns a continuous implicit representation of the dense displacement field. Our method improves the generalization of existing INRs by conditioning the representation on the local neighborhood. Moreover, our approach naturally supports interactive test-time refinement, enhancing its practicality for real-world clinical applications. We visualize the proposed method in \prettyref{fig:method}. In summary, the main contributions of this work are threefold: 1) We introduce a novel implicit framework for medical image registration leveraging learnable kernels. 2) We condition the representation on local keypoint correspondences for improved generalization. 3) We extensively evaluate our method and compare it with several state-of-the-art approaches on zero-shot intra-patient registration tasks. In the following sections, we introduce our method, followed by empirical results and ablation studies supporting the design choices.
\section{Method}

In this work, we cast deformable medical image registration as a signal reconstruction problem. Our goal is to reconstruct a high-dimensional signal $\mathbf{d}: \mathbb{R} ^ 3 \to \mathbb{R}^{3}$ from a sparse set of observations $O = \{ p_i, d_i \}_{i \in \mathbb{N}: 1 \le i \le m}$, where $p_i$ are spatial coordinates $\in \mathbb{R} ^ 3$, and $d_i \in \mathbb{R} ^ 3$ denote the observed signal values. Under this formulation, we evaluate $\mathbf{d}$ on the regular Cartesian grid to represent the displacement field $\phi$ on the voxel domain.
%The signal $\mathbf{d}$ can be viewed as a function that maps a point in the spatial domain $I$ to the associated displacement.

\subsection{Learnable kernels for image registration}

Our key insight is that natural signals, such as displacement fields in medical images, exhibit strong local and global structure that can be encoded in a learnable basis representation. We adopt a kernelised formulation and compute $\mathbf{d}(x)$ as
\begin{equation}
\label{eq:kernel}
    \mathbf{d}(x) = \sum_{y \in \mathcal{N}(x)} \mathbf{w}(x, y) \cdot \mathbf{d}(y),
\end{equation} where $\mathcal{N}(x)$ represents the neighborhood of $x$, and the kernel $\mathbf{w}$ determines the contribution of each neighbor to the final displacement at $x$. The signal observations can be obtained via point-to-point correspondences between the moving and fixed images, as described later on. While $\mathbf{w}$ is defined a priori in traditional techniques, in our method it is parameterized with a neural network. In particular, we design a dual-stream attention mechanism \cite{attention_transformer} to disentangle geometric and semantic information through two dedicated attention heads $\mathcal{H}_s, \mathcal{H}_f$, respectively:
\begin{equation}
\label{eq:attn}
    a(x, y) = \mathcal{H}_s(\mathcal{E}_s(x), \mathcal{E}_s({y}))) + \mathcal{H}_f(\mathcal{E}_f(x_f) + \mathcal{E}_f({y_f}))) + b(x, y), \quad y \in \mathcal{N}(x).
\end{equation}
$a(x, y)$ encodes the attention score between two points in the image domain $x, y$, and $b(x, y) = \frac{1}{1 + \| x - y \|^2}$ is the spatial bias component. $\mathcal{E}_s$ and $\mathcal{E}_f$ are the geometric and semantic feature encoders respectively, parameterized with a three-layer MLP with $128$ hidden units and ReLU activations; overall our learnable kernel module consists of $\approx 150$ thousands trainable parameters.
We extract dense semantic features from the inputs with a learnable encoder $\mathcal{F}$, which are sampled at $x, y$ to obtain the feature vectors $x_f, y_f$ respectively; $\mathcal{F}$ is parameterized with a multi-scale UNet encoder \cite{unet}. To calculate $\mathbf{w}$ we apply the softmax operation on the attention scores in the neighborhood $\mathbf{w}(x, y) = \frac{e^{a(x, y)}}{\sum_{y \in \mathcal{N}(x)} e^{a(x, y)}}$.

\subsection{Conditioning}

To improve consistency and generalization, we propose to condition the reconstruction of the displacement field on the local neighborhood of displacements. To do so, we first detect a set of salient points on the image, and then estimate the optimal correspondences in the other image using a differentiable cost-volume layer on the learned multi-scale features \cite{flownet_costvolume, pdd_net}. These correspondences are the signal observations $\mathbf{d}(y), y \in \mathcal{N}(x)$, used in the reconstruction of the displacement field. To construct the neighborhood $\mathcal{N}(x)$ of the query point $x$, we include the nearest $K = 30$ keypoints and associated correspondences.
In our experiments, we find that traditional detectors such as F{\"o}rstner \cite{iterlbp} and SIFT \cite{sift}, as well as deep-learning ones such as DISK \cite{disk_points} and SuperPoint \cite{superpoint}, provide a sufficient number of keypoints to achieve competitive performance, with minimal computation overhead. When using SIFT, DISK, and SuperPoint detectors we process each $2$D slice in the input volume separately and retain only the points with confidence greater than a pre-defined tolerance. During training, we limit the number of keypoints to $1024$ by farthest point sampling.\\

\minisection{Test-time Interactive Refinement} A unique feature of our method is the support for efficient test-time interactive refinement. We leverage the implicit kernelised representation and compute localized updates to $\mathbf{w}(x, u)$, where $u \in \mathcal{N}(x)$ and is a new point-to-point correspondence between images, without requiring an additional full registration.

\subsection{Training} 

Following related work, we predict the final displacement field incrementally with a multi-scale pipeline \cite{nicetrans, modet, winet}. We reconstruct the optimal dense deformation field at each scale, and use this estimate to warp the features at the next scale, iterating until full-resolution; in our experiments, we use five scales.
%At the coarsest scale, the receptive field encompasses large anatomical regions, facilitating global alignment, while at finer scales detailed structural information is preserved necessary for precise registration.
We train our method with usual registration loss functions, imposing the normalized cross-correlation (NCC) loss to promote image similarity, and the isotropic diffusion as regularizer \cite{VoxelMorph}. If available, we also warp the segmentation masks, and impose the Dice loss \cite{transmorph} on the warped and fixed segmentation masks. Similarly, if landmarks are available, we include the Euclidean distance between the fixed and warped landmarks in the loss function. We use the same weight for each loss function. Source code will be released upon acceptance. %is available anonymously at \url{https://anonymous.4open.science/r/6F91/model.py}.
\section{Experiments and Results}
\label{sec:results}

We compare our method with state-of-the-art registration approaches, including six incremental multi-scale registration techniques (corrMLP \cite{corrmlp}, H-ViT \cite{hvit}, ModeT \cite{modet}, NICE-Trans \cite{nicetrans}, RDP \cite{rdp}, WiNet \cite{winet}), the implicit method AM SIREN \cite{towards_gener_implicit_reg}, and the recent foundation model uniGradICON \cite{unigradicon}. To ensure a fair comparison, we fine-tuned uniGradICON with $50$ test-time iterations ($\approx 1$ minute). We implemented a multi-scale version of the ``3l-512'' AM SIREN architecture, totaling $7.8$ millions of trainable parameters. The remaining architectures were configured to have $3.5 {\scriptstyle\pm 0.3}$ millions of trainable parameters by adjusting the number of feature channels. We train all the learning methods until convergence on the validation set (for a maximum of $100$ epochs), and test them using the checkpoint with the best validation metrics; each training run never exceeds $7$ hours. During training we adopt common data augmentation techniques such as Gaussian noise and blurring. We set the same randomization seed in each training run so that each model is trained on exactly the same data.\par
We use two challenging intra-patient registration datasets, namely the public NLST data\footnote{\url{https://www.cancerimagingarchive.net/collection/nlst}} \cite{tcia_2013}, and a large-scale dataset from the \klinik{} containing $96$ colorectal cancer hepatic interventions. In the following tables, we report the $95\%$ confidence intervals in brackets \cite{Confidence_intervals_MICCAI2024}. $^*$ represents statistically significant differences with respect to our results, after Bonferroni correction.\\

\minisection{NLST data} This dataset, featured in the MICCAI Learn2Reg 2023 challenge \cite{learn2reg}, is extensively used to benchmark intra-patient registration methods. The registration task involves estimating lung deformations between inhale and exhale CT scans, a challenging problem due to the presence of large non-linear displacements. In our experiments, we observed that using a mean-squared error (MSE) loss yielded superior results compared to using the NCC loss. To ensure a fair comparison with memory-intensive methods, we resized the input volumes to half of their original resolution ($112 \times 96 \times 112$), and trained all models at this resolution; for the additional pre-processing steps we follow the challenge instructions \cite{learn2reg}. We compare the methods on the salient target registration error (TRE, measured in mm) using the provided ground-truth landmarks, and on the regularity of the deformations, in terms of the standard deviation of the logarithm of the Jacobian determinant (SDlogJ); to evaluate robustness, we also determine the $30$th percentile of largest landmark distances (TRE$30$) \cite{learn2reg}. The results are compiled in \prettyref{tab:nlst}.\begin{table}[!h]
    \centering
    \caption{Quantitative results on the Learn2Reg NLST dataset.}
    
    \begin{tabular}{cccc}
    \hline
    \rowcolor{gray!20} Method\makespace & TRE (mm) $\downarrow$\makespace & TRE$30$ (mm) $\downarrow$\makespace & SDlogJ $\downarrow$\makespace\\\hline

    \textit{ours}\makespace & $1.72 {\scriptstyle\pm 0.43}~[1.40, 2.04]$~\makespace & $1.89 {\scriptstyle\pm 0.38}~[1.60, 2.17]$~\makespace & $0.02$~\makespace \\
    AM SIREN \cite{towards_gener_implicit_reg}\makespace & $3.51 {\scriptstyle\pm 1.06}~[2.71, 4.31]^*$\makespace & $4.11 {\scriptstyle\pm 1.23}~[3.19, 5.04]^*$\makespace & $0.08^*$\makespace \\
    corrMLP \cite{corrmlp}\makespace & $3.30 {\scriptstyle\pm 1.33}~[2.29, 4.30]^*$\makespace & $3.80 {\scriptstyle\pm 1.42}~[2.73, 4.87]^*$\makespace & $0.05^*$\makespace \\
    H-ViT \cite{hvit}\makespace & $3.77 {\scriptstyle\pm 1.39}~[2.72, 4.82]^*$\makespace & $4.45 {\scriptstyle\pm 1.57}~[3.26, 5.63]^*$\makespace & $0.05^*$\makespace \\
    ModeT \cite{modet}\makespace & $2.33 {\scriptstyle\pm 0.76}~[1.75, 2.90]$~\makespace & $2.51 {\scriptstyle\pm 0.72}~[1.97, 3.05]$~\makespace & $0.06^*$\makespace \\
    NICE-Trans \cite{nicetrans}\makespace & $3.27 {\scriptstyle\pm 1.32}~[2.28, 4.27]^*$\makespace & $3.80 {\scriptstyle\pm 1.67}~[2.54, 5.06]^*$\makespace & $0.07^*$\makespace \\
    RDP \cite{rdp}\makespace & $2.42 {\scriptstyle\pm 1.04}~[1.63, 3.20]$~\makespace & $2.56 {\scriptstyle\pm 1.06}~[1.77, 3.36]$~\makespace & $0.05^*$\makespace \\
    uniGradICON \cite{unigradicon}\makespace & $1.77 {\scriptstyle\pm 0.29}~[1.55, 1.98]$~\makespace & $1.87 {\scriptstyle\pm 0.31}~[1.63, 2.10]$~\makespace & $0.04^*$\makespace \\
    WiNet \cite{winet}\makespace & $3.60 {\scriptstyle\pm 1.31}~[2.61, 4.59]^*$\makespace & $4.17 {\scriptstyle\pm 1.33}~[3.16, 5.17]^*$\makespace & $0.03^*$\makespace \\
    \hline

    \end{tabular}
    
    \label{tab:nlst}
\end{table}

\minisection{Colorectal liver cancer data} This dataset includes $96$ CT scans from different patients before and immediately after radio-frequency ablation \cite{bale_stereotactic_2011_real}, acquired in arterial and venous phase respectively. This minimally invasive intervention induces highly non-linear deformations in the liver, primarily due to respiratory motion and tissue shrinkage \cite{liver_motion_caused_by_ventilation, local_biomek_tissue_shrinkage_mwa}, as well as significant intensity changes near the tumor region. These challenges make the dataset particularly valuable for evaluating the robustness and accuracy of registration methods under complex, real-world conditions. Ground-truth segmentation masks for liver, tumor, and treatment area have been semi-automatically obtained, checked, and corrected by two clinicians at the \klinik{}. Pre-processing steps involve resampling the volumes to the same voxel spacing ($3.0 \times 1.4 \times 1.4$ mm$^3$), cropping a region of $[80 \times 192 \times 192]$ voxels around the liver mask obtained with TotalSegmentator \cite{total_segmentator_ct}, masking the image intensities using the $5$th and $95$th percentiles, and normalizing them to $[0, 1]$. We used a three cross-fold training scheme, with $48$ cases for training, $16$ for validation, and $32$ for testing. We compare with related work on the following two tasks: liver registration accuracy and safety margin assessment (SMA). For the liver registration, we compute the average symmetric surface distance (ASSD), and the $95$th percentile of the Hausdorff distance (HD95) on the liver mask, both measured in mm. To assess the reliability of the deformations, we compare the methods on SMA, a critical task in minimally invasive workflows, necessary to evaluate the treatment success and to prevent local recurrence. Following related work \cite{identification_a0}, we measure the distance between the warped treatment area and the pre-operative tumor, using a $5$ mm margin as the threshold to determine treatment success \cite{mam_gregor}: this is widely recognized as an independent predictor of local tumor recurrence \cite{multicenter_intersoftware_valid}. We calculate the receiver operating characteristic (ROC) curve for this classification task \cite{identification_a0}, and report the results in \prettyref{tab:klinik}. In the last row of the table, we show the results from a state-of-the-art commercial software, specifically designed for this task \cite{ablationfit}. For comprehensive visual results, we refer to the supplementary material. \begin{table}[!h]
    \centering
    \caption{Quantitative results on the dataset from the \klinik{}.}
    
    \begin{tabular}{cccc}
    \hline
    \rowcolor{gray!20} Method\makespace & Liver ASSD (mm) $\downarrow$\makespace & Liver HD95 (mm) $\downarrow$\makespace & SMA (\%) $\uparrow$\\\hline

    \textit{ours}\makespace & $1.10 {\scriptstyle\pm 0.96}~[0.90, 1.29]$~\makespace & $4.71 {\scriptstyle\pm 3.78}~[3.94, 5.48]$~\makespace & $70.59$\makespace\\
    AM SIREN \cite{towards_gener_implicit_reg}\makespace & $0.99 {\scriptstyle\pm 0.90}~[0.81, 1.17]$~\makespace & $4.86 {\scriptstyle\pm 3.76}~[4.10, 5.63]$~\makespace & $52.94$\makespace \\
    corrMLP \cite{corrmlp}\makespace & $2.53 {\scriptstyle\pm 2.46}~[2.03, 3.04]^*$\makespace & $8.51 {\scriptstyle\pm 6.98}~[7.09, 9.93]^*$\makespace & $53.27$\makespace \\
    H-ViT \cite{hvit}\makespace & $1.63 {\scriptstyle\pm 1.24}~[1.37, 1.88]^*$\makespace & $6.20 {\scriptstyle\pm 4.48}~[5.29, 7.11]$~\makespace & $50.33$\makespace \\
    ModeT \cite{modet}\makespace & $0.98 {\scriptstyle\pm 0.87}~[0.80, 1.15]$~\makespace & $4.77 {\scriptstyle\pm 3.72}~[4.02, 5.53]$~\makespace & $56.21$\makespace \\
    NICE-Trans \cite{nicetrans}\makespace & $1.26 {\scriptstyle\pm 1.13}~[1.03, 1.49]$~\makespace & $5.58 {\scriptstyle\pm 4.18}~[4.73, 6.43]$~\makespace & $59.15$\makespace \\
    RDP \cite{rdp}\makespace & $0.82 {\scriptstyle\pm 0.88}~[0.64, 1.00]$~\makespace & $4.13 {\scriptstyle\pm 3.45}~[3.42, 4.83]$~\makespace & $58.82$\makespace \\
    uniGradICON \cite{unigradicon}\makespace & $1.01 {\scriptstyle\pm 0.88}~[0.84, 1.19]$~\makespace & $4.29 {\scriptstyle\pm 3.15}~[3.65, 4.93]$~\makespace & $64.71$\makespace \\
    WiNet \cite{winet}\makespace & $1.68 {\scriptstyle\pm 1.27}~[1.42, 1.94]^*$\makespace & $6.33 {\scriptstyle\pm 4.61}~[5.39, 7.26]$~\makespace & $50.65$\makespace \\\hline
    Ablation-fit \cite{ablationfit}\makespace & - & - & $71.24$\makespace \\
    \hline

    \end{tabular}
    
    \label{tab:klinik}
\end{table}

\section{Discussion, Limitations, and Conclusion}
\label{sec:discussion}

The empirical results on the public NLST dataset demonstrate that our method either outperforms or pars state-of-the-art registration accuracy, while generating smoother deformations. In the case of the internal colorectal cancer dataset, while the liver surface is accurately aligned by most approaches, safety margin assessment remains a significant challenge. Notably, ours is the only AI method matching the performance of specialized commercial systems for this task, with dramatically reduced inference time ($1.6 {\scriptstyle\pm 0.3}$ seconds compared to over two minutes), underscoring its potential for clinical translation.\\

\minisection{Ablation studies} To analyze the factors influencing the performance of our method, we conduct ablation studies on the core architectural elements using the public NLST dataset for reproducibility. The results, summarized in \prettyref{tab:ablations}, highlight the importance of each component.
In particular, the proposed local conditioning offers a key advantage in terms of deformation accuracy and regularity, with either attention head. While thin-plate splines (TPS) extrapolation yields higher performance than baseline, the difference is not statistically significant, highlighting the need for a learnable extrapolation mechanism. Further, we observe that keypoint detections from deep-learning methods (DISK \cite{disk_points}, SuperPoint \cite{superpoint}) remain effective even under significant domain shift.\\

\minisection{Limitations} We note that our current implementation allocates $17.8 {\scriptstyle\pm 1.4}$ GB of GPU memory during training due to the dense voxel-wise sampling and cost-volume computations. We are actively developing more efficient implementations to enhance scalability, especially for resource-constrained clinical environments.\begin{table}[!h]
    \centering
    \caption{Ablation studies on the NLST data. $^*$ represents statistically significant differences with respect to the baseline, the multi-scale implementation of the “3l-512” AM SIREN \cite{towards_gener_implicit_reg}, after Bonferroni correction. We use the following abbreviations: \textit{kpts.} for keypoints, and \textit{extrap.} for extrapolation.}
    
    \begin{tabular}{cccc}
    \hline
    \rowcolor{gray!20} Method\makespace & TRE (mm) $\downarrow$\makespace & TRE$30$ (mm) $\downarrow$\makespace & SDlogJ $\downarrow$\makespace\\\hline
    
    baseline\makespace & $3.51 {\scriptstyle\pm 1.06}~[2.71, 4.31]$~\makespace & $4.11 {\scriptstyle\pm 1.23}~[3.19, 5.04]~$\makespace & $0.08~$\makespace\\

    %\textit{ours}\makespace & $1.72 {\scriptstyle\pm 0.43}~[1.40, 2.04]^*$\makespace & $1.89 {\scriptstyle\pm 0.38}~[1.60, 2.17]^*$\makespace & $0.02^*$\makespace \\
    only $\mathcal{H}_s$\makespace & $2.00 {\scriptstyle\pm 0.54}~[1.59, 2.40]^*$\makespace & $2.24 {\scriptstyle\pm 0.60}~[1.79, 2.69]^*$\makespace & $0.01^*$\makespace\\
    only $\mathcal{H}_f$\makespace & $1.79 {\scriptstyle\pm 0.45}~[1.45, 2.14]^*$\makespace & $1.98 {\scriptstyle\pm 0.41}~[1.67, 2.29]^*$\makespace & $0.01^*$\makespace \\

    TPS extrap.\makespace & $2.23 {\scriptstyle\pm 0.62}~[1.76, 2.69]$~\makespace & $2.53 {\scriptstyle\pm 0.67}~[2.02, 3.04]^*$\makespace & $0.02^*$\makespace \\
    DISK  kpts. \makespace & $1.75 {\scriptstyle\pm 0.35}~[1.48, 2.01]^*$\makespace & $1.89 {\scriptstyle\pm 0.36}~[1.62, 2.17]^*$\makespace & $0.02^*$\makespace \\
    SIFT  kpts. \makespace & $1.78 {\scriptstyle\pm 0.45}~[1.43, 2.12]^*$\makespace & $1.95 {\scriptstyle\pm 0.52}~[1.55, 2.34]^*$\makespace & $0.02^*$\makespace \\
    SuperPoint kpts. \makespace & $2.15 {\scriptstyle\pm 0.64}~[1.66, 2.63]^*$\makespace & $2.42 {\scriptstyle\pm 0.66}~[1.92, 2.91]^*$\makespace & $0.02^*$\makespace \\\hline

    \end{tabular}
    
    \label{tab:ablations}
\end{table}

\begin{figure}[!h]
\centering

\begin{subfigure}[b]{\textwidth}
    \centering
    \includegraphics[width=\textwidth]{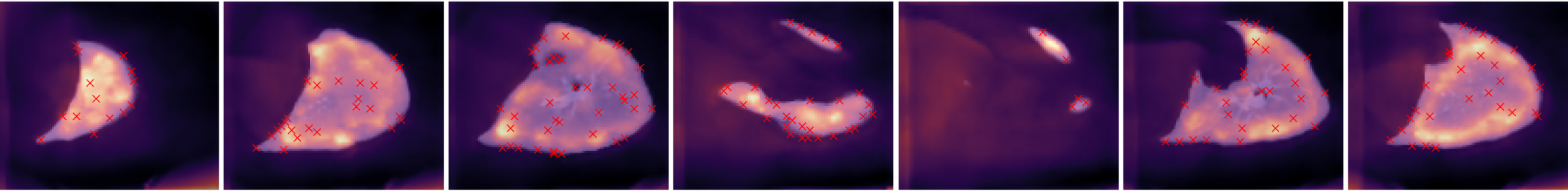}
    \caption{Case $101$ from the NLST data, landmarks TRE $=1.10$ mm.}
    \label{fig:confidence_nlst}
\end{subfigure}

\begin{subfigure}[b]{\textwidth}
    \centering
    \includegraphics[width=\textwidth]{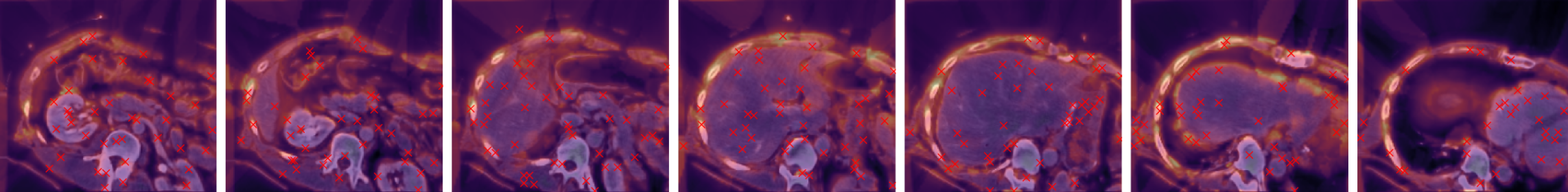}
    \caption{Case $130$ from the \klinik{}, liver ASSD $=0.87$ mm.}
    \label{fig:confidence_klinik}
\end{subfigure}
\caption{Overlay of the prediction ``confidence'' on the warped volume, with detected keypoints marked in red.}

\label{fig:user}
\end{figure}

\minisection{Test-time behavior} Finally, we highlight the practicality of our approach by computing the standard deviation of the attention scores as a proxy for the ``confidence'' of the predicted displacements \cite{calibration_of_modern_nn}, and visualize it in \prettyref{fig:user}. Then, we examine the value of interactive refinement by re-computing the displacements based on the provided ground-truth landmarks: we achieve a TRE improvement of $5, 9, 13, 14$ \% using $10, 20, 30, 40$ uniformly randomly sampled landmarks respectively.

\minisection{Conclusion and Future Work} In summary, this work introduces a novel implicit framework that achieves a unique balance of accuracy, reliability, and clinical usability by conditioning the signal reconstruction on sparse keypoint correspondences. Our approach not only mitigates generalization issues in existing AI-based registration methods but it also provides a robust and practical solution for real-world clinical applications, achieving performance comparable to specialized commercial systems. In the future, we will conduct further evaluation on different anatomical areas and modalities.
%~\\\input{sections/danksagung}

\newpage  % 8 pages (text, figures, and tables) + up to 2 pages for references only, submit it separately,  Captions should not exceed 100 words.
\bibliographystyle{splncs04}
\bibliography{mybib}

\end{document}